\pdfoutput=1
\documentclass[11pt]{article}

\usepackage{acl}

\usepackage{xcolor}
\usepackage[utf8]{inputenc}
\usepackage{graphicx}
\usepackage[T1]{fontenc}    
\usepackage[english]{babel}
\usepackage{natbib}
\usepackage{xcolor}
\usepackage{booktabs}
\usepackage{hyperref}
\usepackage[htt]{hyphenat} 

\title{A cost--benefit analysis of cross-lingual transfer methods}

\author{Guilherme Moraes Rosa$^{1,2}$ \ \ 
  Luiz Henrique Bonifacio$^{1,2}$\\
  \textbf{Leandro Rodrigues de Souza}$^1$ \ \ 
  \textbf{Roberto Lotufo}$^{1,2}$ \ \ 
  \textbf{Rodrigo Nogueira}$^{1,2,3}$\\
  \\
  $^1$University of Campinas \\
  $^2$NeuralMind \\
  $^3$University of Waterloo
  }

\begin{document}
\maketitle

\begin{abstract}

An effective method for cross-lingual transfer is to fine-tune a bilingual or multilingual model on a supervised dataset in one language and evaluate it on another language in a zero-shot manner.
Translating examples at training time or inference time are also viable alternatives. However, there are costs associated with these methods that are rarely addressed in the literature.
In this work, we analyze cross-lingual methods in terms of their effectiveness (e.g., accuracy), development and deployment costs, as well as their latencies at inference time.
Our experiments on three tasks indicate that the best cross-lingual method is highly task-dependent.
Finally, by combining zero-shot and translation methods, we achieve the state-of-the-art in two datasets used in this work.
Based on these results, we question the need for manually labeled training data in a target language. 
Code and translated datasets are available at \url{https://github.com/unicamp-dl/cross-lingual-analysis}
\end{abstract}





\maketitle

\section{Introduction}

In many languages, a common problem when using machine learning models for natural language processing (NLP) tasks is the low availability of high-quality datasets for fine-tuning~\citep{huang2019matters}, as data annotation is costly both in terms of money and time spent~\citep{ref1, ref2}.
In contrast, multilingual pretrained models show surprisingly good cross-lingual zero-shot performance, i.e., models fine-tuned only on a dataset of a high-resource language, such as English, perform well on another language on the same task~\citep{wu, xlm-r, xue2021mt5}. 
Due to improvements in machine translation in the last few years~\citep{wu2016google,lepikhin2020gshard}, automatically translating datasets from a high-resource to a low-resource language has also become an effective cross-lingual transfer strategy.

Current literature, however, mostly focuses on the development and understanding of transfer learning methods that potentially lead to a better model with respect to some target task metrics, ignoring development costs, such as training data translation, and recurring costs, such as inference cost per example. This work analyzes the feasibility and cost-effectiveness of cross-lingual methods to answer the following question: given the availability of large supervised datasets in English and models pretrained on various languages, what is the most cost-effective way to use these resources for tasks in other languages? 
The answer to this question allows us to effectively develop and deploy natural language processing systems for tasks where there is not sufficient labeled data to fine-tune the models. To answer it, we analyze the following transfer learning techniques: 1) fine-tuning a model on a source language and evaluating it on the target language without translation, i.e., in a zero-shot manner; 2) automatic translation of the training dataset to the target language; 3) automatic translation of the test set to the source language at inference time and evaluation of a model fine-tuned on English.

Our main contribution is to evaluate cross-lingual transfer methods while also considering their financial and computational costs. In addition, we show that automatically translating question answering datasets is not trivial and propose a new method for translating them. Finally, while exploring the best cross-lingual methods, we achieved the state-of-the-art in two datasets in a low-resource language, thus showing that our methodology is sound.

\section{Related Work}


Recent work shows that multilingual models fine-tuned on a specific task perform well on the same task on a different language never seen during fine-tuning~\citep{multibert,artetxe2019,phang2020,cross}. This zero-shot cross-lingual ability allows one to use these models on tasks in languages in which annotated data is scarce. Their effectiveness is quite surprising because they are not generally pretrained in any cross-lingual objective. This behavior fostered several studies that aimed to understand and explore it. For example, \citet{wu} explored the cross-lingual potential of multilingual BERT (mBERT) as a zero-shot language transfer model for NLP tasks such as named-entity recognition (NER) and parsing. They further observed that mBERT performs better in languages that share many subwords. \citet{mbert} have shown that mBERT has good zero-shot cross-lingual transfer performance on NER and POS tagging tasks. \citet{artetxe2019} concluded that neither shared subwords vocabulary nor joint training across multiple languages are necessary to obtain cross-lingual abilities. They have shown that monolingual models are also capable of performing cross-lingual transfer. \citet{cross-mbert} used mBERT to study the impact of linguistic properties of the languages, the architecture of the model, and the learning objectives on the generalization ability of cross-lingual language models. The experiments were conducted in three typologically different languages and they concluded that the lexical overlap between languages contributes little to the cross-lingual success, while the depth of the network plays an important role. 

\citet{cross} proposed two methods to pretrain cross-lingual language models, one unsupervised and the other supervised that achieved state-of-the-art results on cross-lingual classification, unsupervised and supervised machine translation. They also have shown that cross-lingual language models can provide significant improvements on the perplexity of low-resource languages.
\citet{xlm-r} presented a Transformer-based multilingual model named XLM-R, pretrained on one hundred languages and a strong competitor to monolingual models on several zero-shot benchmarks. \citet{cross-ir} showed that multilingual models perform well on cross-lingual document ranking tasks. They also investigated translating the training data and the translation of documents at inference time, and concluded that both approaches achieve competitive results.
\citet{assin-br} analyzed multiple approaches of using mBERT for natural language inference task in Portuguese. They investigated the consequences of adding external data to improve training in two different forms: multilingual data and an automatically translated corpus. They achieved the state-of-the-art on ASSIN corpus using a multilingual pretrained BERT model and showed that using external data did not improve the model's performance or the improvements are not significant.

\citet{deepbrasil} showed that automatically translating examples to English and using a model fine-tuned on an English dataset can outperform multilingual models fine-tuned on the target language.
\citet{isbister2021should} demonstrated that a combination of English language models and modern machine translation outperforms native language models in most Scandinavian languages on sentiment analysis. They argued that it is more effective to translate data from low-resource languages into English than to pretrain a new language model on a low-resource language. Our paper expands this study to the tasks of question answering, natural language inference and passage text ranking.
Both \citet{xue2021mt5} and \citet{goyal2021larger} compared zero-shot and translation of training data approaches, achieving the best results with the latter. We extend the investigation by quantifying the relevant costs embedded during the development and deployment of cross-lingual models.

\section{Methodology}


Below we describe the techniques explored in this work for transferring knowledge in data and models from a source language to a target language.

\subsection{Zero-shot}

Zero-shot cross-lingual transfer refers to the strategy of transferring knowledge learned through datasets and models available in a source language, in which ample resources are often available, to perform tasks in a target language, which typically has fewer labeled data.
One example is to fine-tune mBERT on an English dataset such as SQuAD and directly evaluate it on a question answering dataset in another language such as Portuguese or German.

\subsection{Translate-train}
Typically, a high-quality NLP system is fine-tuned on a large and diverse dataset. However, many of such datasets are only available in a few languages like English and Chinese.
One strategy is to translate these datasets using an automatic translator and fine-tune an NLP model on the translated dataset.
The advantage of this approach is that the translation needs to be done only once, thus there is no extra cost at inference time. Among the disadvantages, there are the cost of translating the entire training dataset, artifacts introduced during translation~\citep{artetxe2020}, and constraints in the input-output format of some tasks such as extractive QA (more on that in Section~\ref{sec:dataset_translation}).

\subsection{Translate-infer}
Due to the existence of several models fine-tuned on high-quality datasets in English, one strategy is to translate the model's input from the target language to English at inference time.
The advantages of this method are the simplicity of implementation and the availability of high-performance machine translation models and off-the-shelf models trained on English tasks. The disadvantages are the possible loss of information due to a noisy translation, the cost of translation and longer inference time, as the latency of the translation model will be added to the latency of the entire system.

\section{Experiments}
\subsection{Datasets}
We evaluate the cross-lingual methods on question answering (QA), natural language inference (NLI) and passage ranking tasks. We use FaQuAD and ASSIN2 for experiments in Portuguese, GermanQuAD and XNLI for experiments in German, ViQuAD and XNLI for experiments in Vietnamese and MS MARCO for passage ranking experiments in all three languages. The models are fine-tuned on SQuAD, MNLI and MS MARCO.



\textbf{SQuAD:} The Stanford Question Answering Dataset (SQuAD) \citep{squad2016} is a question answering dataset whose objective is, given a question and a context, to predict the answer as a context span. The dataset is formed by 107,785 question, context and answer triples manually annotated. From this, 80\% of the examples were destined for the training set, 10\% for the development set and 10\% for the test set.


\textbf{FaQuAD:} \citet{faquad} introduced FaQuAD, a dataset of reading comprehension in Portuguese, whose domain is about Brazilian higher education institutions.
The dataset follows the format of SQuAD and consists of 837 questions for training and 63 for testing, covering 249 paragraphs taken from 18 official documents from a computer science college at a Brazilian federal university and 21 Wikipedia articles related to the Brazilian higher education system.
The objective is the same as SQuAD, i.e., to predict an answer span given a question and a context as input.
Its main metrics are the exact match and token-level F1, i.e., the prediction and ground truth answers are treated as bags of tokens and then their F1 is computed.

\textbf{GermanQuAD:} GermanQuAD \citep{germanquad2021} is a mannually annotated dataset of 13,722 extractive question/answer pairs taken from German Wikipedia. The dataset has 11,518 examples for training, 2,204 examples for testing and follows SQuAD data format, objective and metrics.

\textbf{ViQuAD:} Vietnamese Question Answering Dataset \citep{viquad} is a dataset to evaluate machine reading comprehension models. The dataset is made of 23,074 human-generated question-answer pairs based on 5,109 passages of 174 Vietnamese articles from Wikipedia, in which 18,579 examples are used for training, 2,285 for development and 2,210 for testing.

\textbf{MNLI:} The Multi-Genre Natural Language Inference (MNLI) \citep{mnli} is a collection of almost 433,000 pairs of sentences with annotations of entailment, divided into 392,702 training examples, 20,000 for evaluation and 20,000 for testing. 
Given a premise sentence and a hypothesis sentence, the task is to predict whether the premise entails the hypothesis, contradicts the hypothesis or neither of both. The corpus consists of sentences derived from ten different sources, reflecting ten different genres of written and spoken English, which include transcribed speech, letters, fiction and government reports.



\textbf{ASSIN2:} Avaliação de Similaridade Semântica e Inferência Textual (ASSIN2) \citep{ASSIN2} is the second edition of ASSIN, a shared task in Portuguese that evaluates two types of relationships between sentences: semantic textual similarity, that consists of quantifying the level of semantic equivalence between two sentences, and textual entailment recognition (which we refer as natural language inference in this paper), whose objective is to classify whether a first sentence entails the second.
The ASSIN2 dataset has about 10,000 pairs of sentences, 6,500 of which are used for training, 500 for validation and 2,448 for testing. 
The main metric to evaluate the performance on ASSIN2 is accuracy.

\textbf{XNLI:} XNLI \citep{xnli} is an evaluation set for cross-lingual language understanding by extending the development and test sets of the MNLI to 15 languages, including German and Vietnamese. We use accuracy to evaluate our models.

\textbf{MS MARCO \& mMARCO:} The MS MARCO passage ranking~\citep{marco} is a large-scale dataset comprising 8.8M passages obtained from the top 10 results retrieved from the Bing search engine using 1M queries.
The training set contains about 530,000 pairs of query-relevant passages. The development and test sets contain about 6,900 queries each. The test set annotations are kept hidden, and a public submission to the leaderboard is necessary to evaluate the model effectiveness.
The mMARCO dataset \cite{DBLP:journals/corr/abs-2108-13897} is a multilingual version derived from the MS MARCO dataset. The original dataset was translated into 8 different languages. All the dataset features such as the number of passages, queries and relevant query-passages pairs were preserved during translation. 
The mean reciprocal rank of the top 10 passages (MRR@10) is the official metric for this benchmark.

\subsection{Dataset Translation}
\label{sec:dataset_translation}

In this work, we compare two translation approaches to evaluate our models. In the first approach, which we refer to as \textit{open-source}, we use translation models from \citet{TiedemannThottingal:EAMT2020}, which were trained using the Marian-NMT framework~\citep{mariannmt}. Marian-NMT is a neural machine translation framework originally written in C++ for fast training and translation. We use the following models available at HuggingFace~\citep{hf}:  \texttt{Helsinki-NLP/opus-mt-en-ROMANCE} for translating from English to Portuguese,
\texttt{Helsinki-NLP/opus-mt-en-de}  for translating from English to German and
\texttt{Helsinki-NLP/opus-mt-en-vi}  for translating from English to Vietnamese.
In the second approach, which we refer to as \textit{commercial}, we use the Google Translate API.

Each task has its own input-output format, text size and annotation style. Thus, we use a different translation method for each task, as described below.

\begin{figure*}[h]
  \centering
  \includegraphics[width=14cm]{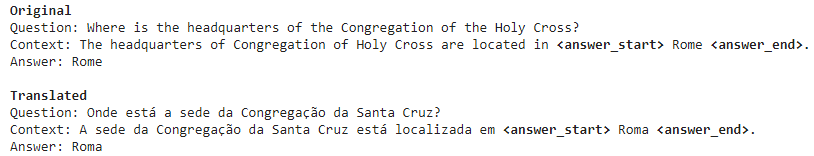}
  \caption{An example of our proposed translation method for extractive QA datasets.}
  \label{fig:nq}
\end{figure*}

\textbf{QA datasets:} We adapted the translation approach used to create the XQuAD dataset~\citep{artetxe2019}, in which human translators were asked to translate the context keeping special symbols inserted to mark the answer span. We translate both context and questions following a similar procedure, but in an automatic manner. To ensure that the final answer is contained within the context, we do not translate the answer separately. Instead, we mark its beginning and end in the context paragraph using special delimiter symbols (e.g. \texttt{<answer\_start>} and \texttt{<answer\_end>}). We then translate the context expecting the model to keep these symbols in the correct positions during translation. We finally extract the answer and the position of the answer span from the translated text based on these delimiters. See Figure~\ref{fig:nq} for an example.

However, this strategy does not always work for open-source models. In some examples, at least one of the delimiters was not kept in the translated context. To address this problem, we fine-tune the translation model on examples that include these delimiters.
We noticed that the open-source models translate single sentences better than multiple sentences. Thus, we translate each sentence in the context independently. Translation using a commercial API is faster and less complicated. We simply translate the entire context at once using the strategy explained above.

Using a batch size equal to the number of sentences in one context, the translation process for SQuAD takes around 34 hours using an open source model and 6 hours using a commercial API, while translating GermanQuAD and ViQuAD take around 23 minutes each for the open source model and 7 minutes using a commercial API. FaQuAD, due to its reduced size, takes 1.5 minutes and 15 seconds, respectively. 

\textbf{NLI datasets:} NLI datasets are much easier to translate since examples are made of only two sentences, which are translated independently. Thus, there are no special delimiters to consider. The resulting translations are concatenated to form the translated example.
\citet{artetxe2020} argue that translating the premise and the hypothesis separately could affect the generalization ability of the models, introducing noise into the dataset. Translating both sentences together would provide more context to the translation model, thus improving the translation quality. However, due to the difficulties mentioned above in finding delimiters of a translated text, we did not experiment with this approach. 
The translation of the MNLI dataset with an open source model takes about 2 hours and 15 minutes, while using a commercial API takes almost 2 hours, both with a batch size of 32. Meanwhile, the translation of ASSIN2 and XNLI takes less than 1 minute.

\textbf{Passage ranking datasets:} The translation of the MS MARCO dataset covers 8 different languages so far. The translation process was conducted equally for all selected languages and it is described in details on mMARCO work \cite{DBLP:journals/corr/abs-2108-13897}.
In this work, we use three different languages from mMARCO: German, Portuguese and Vietnamese. We name the original translation process as \textit{translate-train} approach.

We explore other two translation strategies, named \textit{Strategy 1} and \textit{Strategy 2}, which are translations at inference time. Strategy 1 consists of translating the whole dataset before reranking and, during inference, translating only the queries. This approach is preferred for search systems that receive many queries but retrieve from a small collection.
The second strategy is to translate queries and the top 1000 passages returned before feeding them to the reranker. This strategy is useful when the search system retrieves from a large collection, but it is expected to process a limited number of queries during its lifetime.
The open source translation of the 8.8M passages takes approximately 50 hours on a V100 GPU using batches of size 32. On the other hand, translating using a commercial API takes about 30 hours.

\subsection{Models}

In this section, we describe the models used in this work.

\textbf{BERT-en:}
Bidirectional Encoder Representations from Transformers (BERT) is an open-sourced pretrained language model based on the Transformer's architecture \citep{vaswani2017attention}. We use its base version in all experiments.

\textbf{BERT-pt:}
BERTimbau \citep{bertimbau} is a BERT model pretrained on a Brazilian Portuguese corpus. BERTimbau improved the state-of-the-art on tasks like semantic textual similarity, natural language inference and named entity recognition in Portuguese. 
The model was pretrained on brWaC \citep{brwac}, a corpus containing 2.68 billion tokens from 3.53 million documents. BERTimbau was provided in two sizes, Base and Large.
We call BERTimbau as BERT-pt to easily distinguish it from BERT-en.

\textbf{mBERT:}
mBERT \citep{devlin2018bert} is a BERT model pretrained using the Masked Language Model objective on the Wikipedia articles of 104 languages with a shared word piece vocabulary. The model follows the same architecture of BERT.



\textbf{mT5:} 
Multilingual T5 or mT5 \citep{xue2021mt5} is a multilingual variant of T5 that was pretrained on the mC4 dataset which contains 101 languages. The model architecture and training procedure is based on T5 with small differences, i.e., the increase in the number of parameters that comes from a larger vocabulary to handle all languages.

\subsection{Training and Inference}

\textbf{QA and NLI:}
In our QA and NLI experiments, we use BERT-pt, BERT-en and mBERT models from the HuggingFace library~\citep{hf} fine-tuned on SQuAD and MNLI. Our QA models were fine-tuned on a single Tesla V100 GPU with a constant learning rate of 2e-5, over five epochs, with Adam~\citep{adam} optimizer, batch size of 12 and a maximum of 384 input tokens. Our NLI models were fine-tuned with a V100, learning rate of 2e-5 over three epochs, batch size of 32 and using Adam optimizer with an input of 128 tokens. 

We experiment with two variants of each training set to fine-tune our BERT models: the originals in English and the machine-translated versions in Portuguese, German and Vietnamese. We do the same to evaluate each test set, but, in this case, the originals are in Portuguese, German and Vietnamese and the machine-translated versions are in English. In summary, we try four different approaches: 1) Fine-tune BERT-pt or mBERT on English and evaluate on the target language (zero-shot); 2) Fine-tune BERT-pt or mBERT on a target-translated training set and evaluate it on the same target language (translate-train); 3) Fine-tune BERT-en on SQuAD or MNLI and evaluate it on the test set translated to English (translate-infer BERT-en); 4) As a control experiment, fine-tune BERT-pt or mBERT in English and evaluate it on the test set translated to English (translate-infer BERT-target);

\textbf{Passage ranking:}
We fine-tune mT5 on the passage ranking task following a similar procedure proposed by \citet{nogueira2020document}. The input sequence is formed by a query-passage pair and the model is fine-tuned to return the tokens ``yes'' or ``no'', for German, Portuguese and Vietnamese languages.

All our fine-tuned models are evaluated on the development set of a translated version of MS MARCO. Also, in order to explore the zero-shot performance of mT5, we fine-tune this model on English MS MARCO dataset and evaluated on the three considered languages.


As proposed by \citep{xue2021mt5}, we also fine-tune mT5 on a bilingual version of MS MARCO, i.e., a dataset formed by joining the original version of MS MARCO in English with a translated version in Portuguese, German or Vietnamese. 
In all settings, we fine-tune a mT5-base for 100k steps using batches of size 128, a learning rate of $10^{-3}$ and the AdaFactor optimizer~\citep{shazeer2018adafactor}.
We use a Google's TPU v3-8 to fine-tune and evaluate. While fine-tuning takes about 12 hours, the inference time is approximately 5 hours.
We also established a baseline using BM25 implemented by Anserini~\citep{10.1145/3077136.3080721}. Besides providing baseline results, BM25 also serves as the initial retrieval module for the mT5 reranking models. The top 1000 passages ranked according to BM25 are used as candidate passages to the rerankers.

\begin{table*}[ht]

\centering\centering\resizebox{1.0\textwidth}{!}{
 \begin{tabular}{l|c|c|c|c|c|c} 
 \toprule
 \textbf{Method} & \multicolumn{3}{c|}{\textbf{Score}} &  \multicolumn{1}{c|}{\textbf{One-time Cost}} & \multicolumn{1}{c|}{\textbf{Recurring Cost}} & \textbf{Added latency} \\
 & PT & DE & VI & \multicolumn{1}{c|}{\textbf{(USD)}} & \multicolumn{1}{c|}{\textbf{(USD/1k ex.)}} & \textbf{(s/batch)}\\
 \midrule
 \textbf{QA} & \multicolumn{3}{c|}{\textbf{F1}} & & & \\
 Zero-shot &  0.8240 & 0.6497 & 0.6472 & - & - & -  \\ 
 Translate-train (Open S) & 0.7391 & 0.4602 & 0.4840 & 2.77 & - & -  \\
 Translate-train (Comm) & 0.7334 & 0.6478 & 0.6109 & 299.17 & - & -  \\
 Translate-infer (Open S) & 0.6772 & 0.5868 & 0.4198 & - & 0.06 & 2.50 \\
 Translate-infer (Comm) & 0.6096 & 0.5887 & 0.4200 & - & 16.78 & 0.23 \\
 \midrule
 \textbf{NLI} &  \multicolumn{3}{c|}{\textbf{Accuracy}} & & & \\
 Zero-shot &  0.8291 & 0.7063 & 0.7087 & - & - & -  \\ 
 Translate-train (Open S) & 0.8746 & 0.7564 & 0.7489 & 6.24 & - & -    \\
 Translate-train (Comm) & 0.8921 & 0.7540 & 0.7566 & 941.67 & - & - \\
 Translate-infer (Open S) & 0.8628 & 0.7720 & 0.7209 & - & 0.02 & 0.78\\
 Translate-infer (Comm) & 0.8737 & 0.7846 & 0.7265 & - & 1.50 & 0.76\\
 \midrule
 \textbf{Ranking} &  \multicolumn{3}{c|}{\textbf{MRR@10}} & & &\\
 Zero-shot &  0.2930 & 0.2796 & 0.2414 & - & - & - \\ 
 Translate-train (Open S) & 0.2850 & 0.2640 & - & 141.27 & - & - \\
 Translate-train (Comm) & 0.3020 & 0.2917 & 0.2648 & 50,793.00 & - & - \\
 Translate-infer - Strategy 1 (Open S)& \multicolumn{3}{c|}{0.3810$^{\star}$} & 141.27 & 0.01 & 0.72$^1$ \\
 Translate-infer - Strategy 1 (Comm) & \multicolumn{3}{c|}{0.3810$^{\star}$} & 50,793.00 & 0.70 & 0.72$^1$ \\
 Translate-infer - Strategy 2 (Open S) & \multicolumn{3}{c|}{0.3810$^{\star}$} & - & 16.36 & 680.64$^2$ \\
 Translate-infer - Strategy 2 (Comm) & \multicolumn{3}{c|}{0.3810$^{\star}$} & - & 5,733 & 390.86$^2$ \\
 
 \bottomrule
\end{tabular}
}
\caption{Main results. The symbol $^{\star}$ denotes an upper bound estimation (see text for details). $^1$Each batch is 32 queries, and since passages are already translated, only 32 translations are made. $^2$Each batch is 32 queries, each paired with 1000 passages, totalizing 32k translations.}
\label{table:main_results}
\end{table*} 

\subsection{Translation Costs}

We calculate translation costs for the open-source models and commercial APIs.

\textbf{Open-source:} We use GPUs available on the cloud to translate using open-source models. As of December 2021, the cost of a V100 GPU at Google Cloud was 2.48 USD per hour and at IBM Cloud was 3.06 USD per hour. We use their average of prices to calculate the cost of translating the training set (one-time cost) and 1,000 test examples (recurring cost) using a batch of 32 examples.

\textbf{Commercial:} Based on the average value of translate APIs from Google, IBM and Microsoft, which at the time of this publication are 20 USD per million of characters translated for Google and IBM and 10 USD per million of characters translated for Microsoft, we calculate the amount that would have been spent if we had used them.
In the appendix, we present in Tables~\ref{table:qa_stats}, \ref{table:nli_stats} and \ref{table:msmarco_stats} relevant dataset statistics used to report the costs in Section~\ref{sec:results}.

\textbf{Added latency:} We also compute the added latency for translate-infer, which is the average time spent to translate a single batch using open-source models on a V100 GPU and it is measured in seconds per batch translated.

\section{Results}
\label{sec:results}

Table~\ref{table:main_results} summarizes our main results considering the performance, costs and latencies involved during the process of fine-tuning and deploying the models.

\subsection{Question Answering Task}

We start by analyzing the results for the question answering task. We perform four main experiments, which are shown in Table~\ref{table:results}.
The zero-shot approach (row 2) outperforms all other strategies in all languages, including achieving at least a competitive performance in two of the three languages (pt and de) when compared to datasets originally created in the target language. It also has the lowest training and inference costs since it does not require any translation.

\begin{table}[ht]
\centering\centering\resizebox{0.50\textwidth}{!}{
 \begin{tabular}{l l | c | c | c | c | c }
 \toprule
 & \textbf{Method} & \textbf{Training data} &\textbf{Model} & \textbf{pt} & \textbf{de} & \textbf{vi}  \\ 
 \midrule
 (1a) & \citet{leandro_2021} & FaQuAD & BERT-pt & 0.5928 & - & - \\
 (1b) & \citet{leandro_2021} & GermanQuAD & BERT-de & - & 0.6863 & - \\
 (1c) & \citet{leandro_2021} & ViQuAD & BERT-vi & - & - & 0.7953 \\
 \midrule
 (2) & Zero-shot & SQuAD-en & BERT-target & 0.8240 & 0.6497 & 0.6472 \\
 
 (3) & Translate-train (Open S) & SQuAD-target & BERT-target & 0.7391 & 0.4602 & 0.4840   \\
 
 (4) & Translate-train (Comm) & SQuAD-target & BERT-target & 0.7334 & 0.6478 & 0.6109   \\
 
 (5) & Translate-infer (Open S) & SQuAD-en & BERT-target & 0.5390 & 0.6318 & 0.4356  \\
 
 (6) & Translate-infer (Comm) & SQuAD-en & BERT-target & 0.6251 & 0.5922 & 0.4439   \\
 
 (7) & Translate-infer (Open S) & SQuAD-en & BERT-en & 0.6772 & 0.5868 & 0.4198  \\
 
 (8) & Translate-infer (Comm) & SQuAD-en & BERT-en & 0.6096 & 0.5887 & 0.4200  \\
 
 \bottomrule
\end{tabular}
}
\caption{\label{tab:results_qa} Results on the QA task.} 
\label{table:results}
\end{table}

The questions and context paragraphs of all QA datasets were translated together to ensure that the answer is present in the context. Thus, the performance of translation methods could, in part, be affected by the translation process since mapping the answer spans across languages is not trivial and introduces some errors.
Considering computational and financial costs, it becomes even more evident that zero-shot is the best approach to this task. Translating using open-source models is cheaper, while using a cloud service generally provides better translation. But even so, both development times are still much longer than the zero-shot approach.

\subsection{Natural Language Inference Task}

This section analyzes the results for the natural language inference task in the same four experiments done before for the QA task. The results are shown in Table~\ref{table:results_nli}.

\begin{table}[ht]
\centering\centering\resizebox{0.50\textwidth}{!}{
 \begin{tabular}{l l | c | c | c| c | c } 
 \toprule
 & \textbf{Method} & \textbf{Training data} & \textbf{Model} & \textbf{pt} & \textbf{de} & \textbf{vi} \\ 
 \midrule
 (1) & \citet{leandro_2021} & ASSIN2-pt & BERT-pt & 0.8656 & - & - \\
 \midrule
 (2) & Zero-shot & MNLI-en & BERT-target & 0.8291 & 0.7063 & 0.7087 \\ 
 
 (3) & Translate-train (Open S) & MNLI-target & BERT-target & 0.8746 & 0.7564 & 0.7489 \\
 
 (4) & Translate-train (Comm) & MNLI-target & BERT-target & 0.8921 & 0.7540 & 0.7566 \\
 
 (5) & Translate-infer (Open S) & MNLI-en & BERT-target & 0.8068 & 0.7662 & 0.7263 \\
 
 (6) & Translate-infer (Comm) & MNLI-en & BERT-target & 0.8059 & 0.7730 & 0.7281 \\
 
 (7) & Translate-infer (Open S) & MNLI-en & BERT-en & 0.8628 & 0.7720 & 0.7209 \\
 
 (8) & Translate-infer (Comm) & MNLI-en & BERT-en & 0.8737 & 0.7846 & 0.7265 \\
 
 \bottomrule
\end{tabular}
}
\caption{\label{tab:results_nli} Results of the NLI task.}
\label{table:results_nli}
\end{table}

\begin{table*}[ht]
\centering\centering\resizebox{1.0\textwidth}{!}{
 \begin{tabular}{l|c|c|c|c} 
 \toprule
 \textbf{Model} & \textbf{Pretrain} &  \textbf{Fine-tune} &  \textbf{F1} & \textbf{Acc}  \\
 \midrule
 mBERT \citep{bertimbau}& 100 languages & ASSIN2-pt &  0.8680 & 0.8680 \\
 IPR \citep{ipr} & 100 languages & ASSIN2-pt &  0.8760 & 0.8760 \\
 Deep Learning Brasil \citep{deepbrasil} & EN &  ASSIN2-en & 0.8830 & 0.8830 \\
 PTT5 \citep{carmo2020ptt5} & EN \& PT & ASSIN2-pt  & 0.8850 & 0.8860 \\
 BERTimbau Large \citep{bertimbau} & EN \& PT &  ASSIN2-pt & 0.9000 & 0.9000 \\
 \midrule
 BERT-pt Base (ours) & EN \& PT & MNLI-en + ASSIN2-pt & 0.8990 & 0.8990 \\
 BERT-pt Large (ours) & EN \& PT & MNLI-en + ASSIN2-pt & 0.9180 & 0.9179 \\
 BERT-pt Large (ours) & EN \& PT & MNLI-pt + ASSIN2-pt &  0.9195 & 0.9195 \\
 BERT-pt Large (ours) & EN \& PT & MNLI-(en+pt) + ASSIN2-pt &  0.9207 & 0.9207 \\
 
 \bottomrule
\end{tabular}
}
\caption{\label{tab:results_assin} Test results on ASSIN2 using the official evaluation script.}
\label{table:results_assin}
\end{table*}

BERT fine-tuned on the translated MNLI (rows 3-4) achieves the best performance in two of the three languages, and it is 2 to 3 points above the BERT-en models fine-tuned on MNLI-en (rows 7-8) and almost up to 5 points above the zero-shot approach (row 2). For German, the best performance is the BERT-en model evaluated on a translated test set (row 8).
In this case, it is not so easy to appoint the best method considering all of our standards of comparison, because in addition to performance, we need to take into account the time and the added cost to translate the training dataset or the translation at inference time. 

Fine-tuning BERT on translated MNLI (rows 3-4) provides the best performance metrics in two of three languages, but as shown in Table~\ref{table:main_results}, translating MNLI could be relatively expensive if using a commercial API. An open-source model is considerably cheaper and takes up to 2 hours to translate almost 400,000 examples. The translate-infer method (rows 5-8) has the advantage of using the original training set in English, but it requires translation to English in every example during inference, causing additional latency to the system. On the other hand, the zero-shot method has the advantage of not requiring translation, but its performance is lower than the two best cross-lingual approaches. In summary, the choice of the best NLI system is highly dependent on the requirements of the application in which the model will be used and the financial resources available for development and deployment. 

Furthermore, all of our four cross-lingual approaches achieve good results compared to the model directly fine-tuned on a dataset originally made in Portuguese (row 1). This shows that, in the absence of datasets in the target language, a cross-lingual approach is an excellent solution to data scarcity.

\subsubsection{Combining Cross-Lingual Methods}

Furthermore, we fine-tune BERT-pt models on different datasets composed of combinations of MNLI and ASSIN2. Both datasets were created for the same task, but they have a different number of classes (MNLI has 3 classes and ASSIN2 has 2). Because of that, we combined the datasets in two ways: (1) We fine-tune using all classes, but during evaluation, we remap predictions of class ``Contradiction'' (present only in MNLI) to the class ``Neutral''; (2) We remap all MNLI's ``Contradiction'' examples to ``Neutral'' and fine-tune on both datasets using only two classes. In preliminary experiments, we found that the second method works better, so we only present results for it. Results in Table~\ref{table:results_assin} show that jointly fine-tuning BERT-pt Large on the English MNLI, MNLI translated to Portuguese and ASSIN2 (in Portuguese) results in the state-of-the-art on ASSIN2.

\subsection{Passage Ranking Task}
Results for the passage ranking task are shown in Table~\ref{tab:msmarco}. All rerankers improve upon the BM25 baseline, thus demonstrating their cross-lingual ability.

\begin{table}[ht]
\centering\centering\resizebox{0.48\textwidth}{!}{
\begin{tabular}{l l | c | c | c| c | c} 
\toprule
& \textbf{Method} & \textbf{Training data} & \textbf{Model} & \textbf{pt} & \textbf{de} & \textbf{vi}\\
\midrule
(1) & \citet{DBLP:journals/corr/abs-2108-13897} & - & BM25 & 0.1410 & 0.1210 & 0.1359   \\
\midrule
(2) & Zero-shot & MS MARCO-en & mT5 & 0.2930 & 0.2796 & 0.2414 \\ 
(3) & Translate-train (Open S) & MS MARCO-target & mT5 & 0.2850 & 0.2640 & - \\ 
(4) & Translate-train (Comm) & MS MARCO-target & mT5 & 0.3020 & 0.2917 & 0.2648 \\ 
(5) & Translate-train (Comm) & MS MARCO-(en+target) & mT5 & 0.3060 & 0.2941 & 0.2689 \\ 
\bottomrule
\end{tabular}
}
\caption{\label{tab:msmarco} Results on the development set of Portuguese MS MARCO.}
\end{table}

While a zero-shot approach is more effective than the BM25 baseline, we can see in Table~\ref{tab:msmarco} that translation improves results when using a commercial API. All models fine-tuned on the API-translated MS MARCO (row 4) perform better than the zero-shot approach (row 2). This suggests that the noise introduced due to translation, in this case, did not harm the learning process. Finally, the results of the model fine-tuned on both original and translated datasets (row 5) outperforms all others by a small margin.

MS MARCO is considerably larger than the other datasets used in this work, resulting in a much higher translation cost, as shown in Table~\ref{table:main_results}. Nevertheless, despite having a lower performance when compared to the commercial API, the translation using an open-source model is almost 360 times cheaper. For translation at inference time (translate-infer), the added latency for strategy 1 is smaller since only queries are translated. On the other hand, the added latency for strategy 2, which translates 1000 passages for each query, is considerably higher. If the collection and the number of queries are both large, then the zero-shot approach is recommended. 

For strategies 1 and 2 of translate-infer, we do not translate back to English the translated versions of MS MARCO since we already have the original in English. Thus, for these methods, we report the same MRR@10 from~\citet{nogueira2020document}, who fine-tuned and evaluated a T5 on the original MS MARCO in English. Note that this result is an upper bound estimation, as the translation would likely introduce artifacts and therefore degrade the quality of the reranker models. To calculate the costs, we also estimate that the dataset translated back to English would have similar statistics (e.g., average passage length) to the original one.
Since MS MARCO evaluation set is automatically translated, these results must be taken cautiously. As shown by \citet{artetxe2020}, translation can add spurious patterns that the model can exploit to obtain better results.



\section{Conclusion}

We investigate methods of transferring knowledge from pretrained models in English to Portuguese, German and Vietnamese for the tasks of question answering, natural language inference and passage ranking, while also analyzing their development and deployment costs. We compare three alternatives: translating the training data, translating test data at inference time, and training and inference without translation, i.e., zero-shot.

Our results reveal that the best approach is highly dependent on the task, the requirements for deployment and the financial resources available.
First, we show that the zero-shot approach works well for all tasks, especially for question answering, whose translation process is more complex than the other two tasks.
Natural language inference datasets are easy and cheap to translate, while passage ranking datasets are more expensive to translate but do not present technical difficulties. We do not recommend the automatic translation of question answering datasets because of the challenges imposed by the task format. However, advances in methods for translating them might result in better cross-lingual question answering models. We also show that translating using Google Translate API generally gives better results than using MarianNMT.

We demonstrate that fine-tuning on both original and translated data provides extra gains for the tasks of natural language inference and passage ranking. With that, we improve the state-of-the-art by 2 points on ASSIN2. On FaQuAD, our zero-shot approach is more than 20 F1 points above the performance of the model fine-tuned on the target data.
Based on those results, we question the common assumption that it is necessary to have labeled training data on the target language to perform well on a task.
Our cross-lingual results suggest that fine-tuning on a large labeled dataset in a foreign language might be enough.


\section*{Acknowledgment}
This research was funded by a grant from Fundação de Amparo à Pesquisa do Estado de São Paulo (FAPESP) 2020/09753-5.

\bibliographystyle{ACM-Reference-Format}
\bibliography{main}

\section{Appendix}

In this section, we provide relevant statistics of the datasets that were used to compute the cost and latency numbers in Table~\ref{table:main_results}.

\begin{table*}

\centering
 \begin{tabular}{l l | c | c | c | c}
 \toprule
 & & \textbf{SQuAD} &\textbf{FaQuAD} & \textbf{GermanQuAD} & \textbf{ViQuAD} \\ 
 \midrule
 (1) & Number of characters (training)  & 17,688,764 & 210,804 & 3,988,977 & 4,287,841 \\
 (2) & Number of characters (test) & 2,068,857 & 38,250 & 1,019,461 & 525,784 \\
 (3) & Avg. chars / example (training) & 936.11 & 999.07  & 346.32 & 230.79 \\
 (4) & Avg. chars / example (test) & 1000.89 & 1006.57 & 462.55 & 230.10 \\
 
 \bottomrule
\end{tabular}
\caption{Statistics of QA datasets.}
\label{table:qa_stats}
\end{table*}

\begin{table*}

\centering
 \begin{tabular}{l l | c | c | c | c}
 \toprule
 & & \textbf{MNLI} & \textbf{ASSIN2} & \textbf{XNLI-de} & \textbf{XNLI-vi}\\ 
 \midrule
 (1) & Number of characters (training) & 56,521,137 & 554,639 & - & - \\
 (2) & Number of characters (test) & 1,379,958 & 219,834 & 778,785 & 658,487 \\
 (3) & Avg. chars / example (training) & 144.49 & 85.33 & - & - \\
 (4) & Avg. chars / example (test) & 140.87 & 89.80 & 155.44 & 131.43 \\
 
 \bottomrule
\end{tabular}
\caption{Statistics of NLI datasets.}
\label{table:nli_stats}
\end{table*}

\begin{table*}

\centering
 \begin{tabular}{l | c | c }
 \toprule
 \textbf{} & \multicolumn{2}{c}{\textbf{MS MARCO}}  \\ 

  & Characters & Chars/example \\
 \midrule
 Collection &  3,047,540,622 & 344.67 \\
 
 Training queries & 28,667,746  & 35.44 \\
 
 Development queries & 3,615,835  & 35.77 \\
 
 \bottomrule
\end{tabular}
\caption{\label{table:msmarco_stats} MS MARCO statistics.}
\end{table*} 


\end{document}